\documentclass[letterpaper]{article} %
\usepackage{arxiv}  %
\usepackage[hyphens]{url}  %
\usepackage{graphicx} %
\usepackage{natbib}  %
\usepackage{caption} %
\usepackage{booktabs}
\usepackage{amsmath}
\usepackage{amssymb}
\usepackage{xcolor}

\title{Query-Only Prompt-Agnostic Attacks on Visual In-Context Learning}
\author{
    Xiao Hu\textsuperscript{1}, Muhammad Ibrahim\textsuperscript{2},
    Hao Wen\textsuperscript{3}, Pin-Yu Chen\textsuperscript{4},\\
    Jianliang Wu\textsuperscript{5},
    Antonio Bianchi\textsuperscript{1}
}
\affiliations{
    \textsuperscript{1}Purdue University\quad
    \textsuperscript{2}Georgia Institute of Technology\\
    \textsuperscript{3}Independent Researcher\quad
    \textsuperscript{4}IBM Research\quad
    \textsuperscript{5}Simon Fraser University\\
    \texttt{\{hu953,antoniob\}@purdue.edu},
    \texttt{wenhao2018@hotmail.com}\\
    \texttt{mibrahim@gatech.edu},
    \texttt{pin-yu.chen@ibm.com},
    \texttt{wujl@sfu.ca}
}

\begin{document}

\maketitle
\pagestyle{plain}
\thispagestyle{plain}

\begin{abstract}
Visual In-Context Learning (VICL) enables promptable segmentation by conditioning RGB predictions on visual demonstrations, making it attractive for auto-labeling, but its robustness under query-only access and unknown deployment prompts remains underexplored. We study query-only, prompt-agnostic targeted attacks where the attacker observes only returned RGB outputs, without access to model internals or deployment prompts. Treating prompt-conditioned RGB rendering as a distinct attack surface, we introduce two appearance-space attacks: Color Hijack (CH) drives foreground colors toward the background to suppress visibility, while Task Hijack (TH) collapses instance-distinguishing colors toward a shared color assignment. To transfer across unseen prompts, we propose a Self-Prompt Surrogate (SPS) constructed from the clean query and its output. In a paired, matched-budget SegGPT comparison, SPS outperforms the Random-Prompt Surrogate (RPS), reducing CH instance foreground retention from \(0.918\) to \(0.155\) and TH palette-aware tolerance distance from \(0.345\) to \(0.225\). Across four VICL models on COCO and HierText, CH suppresses foregrounds across semantic-, instance-, and text-style outputs, while TH collapses instance colors under held-out deployment prompts. After palette-based decoding, TH also reduces the decoded instance count and mask retention while increasing instance merges. Finally, at \(30\%\) corruption, CH-corrupted annotations from a publicly accessible VICL-based auto-labeling pipeline reduce downstream YOLO-seg mAP50 from \(0.761\) to \(0.281\).
\end{abstract}

\section{Introduction}
In-context learning (ICL) enables models to infer tasks from a few demonstrations without updating their parameters~\cite{NEURIPS2020_1457c0d6}. Visual in-context learning (VICL) extends this paradigm to vision by representing demonstrations as image--output pairs that serve as visual prompts~\cite{VisualPrompting2022,Wang_Painter_CVPR,xie2025test}. For segmentation, these prompts specify the task and its RGB output convention, allowing the same model to produce differently encoded segmentation outputs without fine-tuning~\cite{Wang_2023_ICCV}. This makes VICL attractive for auto-labeling, where RGB predictions can be post-processed and exported as annotations for downstream analysis or model training~\cite{autodistill2023seggpt,encord2023seggpt,labellerr2023seggpt}. However, the robustness of VICL-based segmentation under query-only access and unknown deployment prompts remains largely unexplored.

\begin{figure*}[t]
\centering
\includegraphics[width=0.75\textwidth]{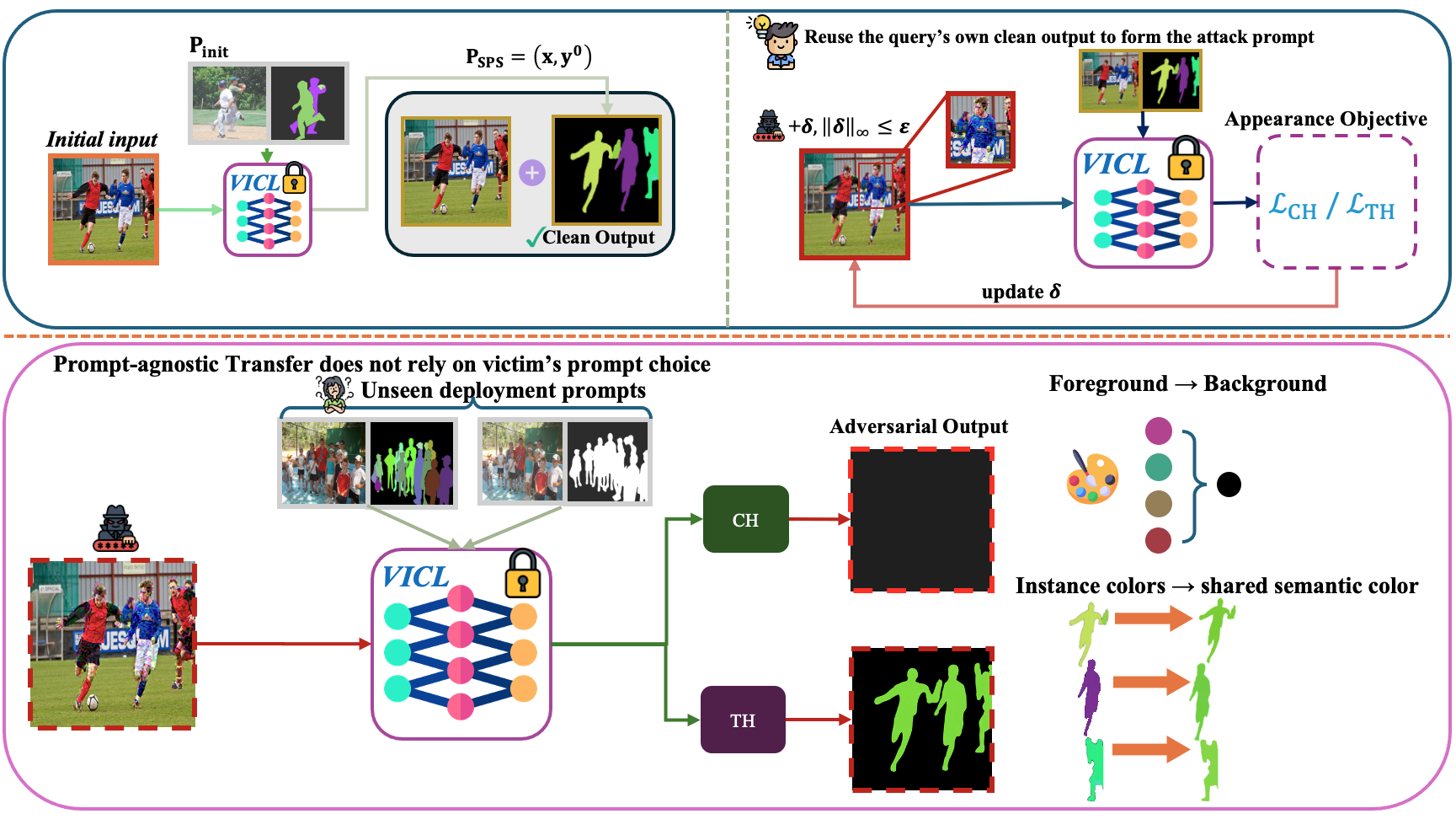}
\caption{Overview of our query-only, prompt-agnostic attacks on VICL-based segmentation. The attack constructs \(P_{\mathrm{SPS}}=(x,y^0)\) from the clean query and its output, then optimizes CH or TH using output-only feedback. Without re-optimization, the adversarial query transfers to unseen deployment prompts: CH suppresses foregrounds, whereas TH collapses instance-specific colors.}
\label{fig:teaser}
\end{figure*}

Existing targeted segmentation attacks are primarily designed for conventional task-specific models~\cite{xie2021segformer,He2017MaskR,Kirillov2019PanopticFP} or SAM-style promptable models~\cite{kirillov2023}. These attacks operate on fixed label or mask spaces~\cite{arnab_cvpr_2018}, including masks conditioned on spatial prompts placed on the query image~\cite{zhang2023attacksamattackingsegmentmodel}, while black-box variants commonly rely on transfer from white-box surrogate models~\cite{A11_xie2017adversarial,cai2023ensemble,10.1109/TMM.2024.3521769}. VICL introduces additional prompt-dependent variability because independent image--output demonstrations jointly specify task semantics, color palettes, and rendering styles. Consequently, an attack optimized for one prompt may overfit to prompt-specific behavior and fail under a different deployment prompt. Prior work also studies data-poisoning backdoors with prompt-dependent behavior~\cite{abad2024contextkeybackdoorattacks}. Constructing a transferable white-box surrogate is also difficult because VICL models differ in their supported tasks, prompt compositions, and visual domains. In our setting, the attacker observes only returned RGB predictions and has no access to model internals or the prompt used at deployment. Existing label- or mask-space transfer attacks therefore do not directly address targeted manipulation of prompt-conditioned RGB outputs under an unknown deployment prompt.

Motivated by this gap, we study query-only targeted attacks on VICL-based segmentation under a prompt-agnostic threat model. During attack construction, the attacker queries the target system using attacker-controlled visual demonstrations and observes only its RGB outputs. The resulting adversarial image is later processed by the same system under an independently selected deployment prompt that is neither observed nor controlled by the attacker. The goal is to perturb only the query image so that its segmentation exhibits an attacker-defined appearance-level effect that transfers across unseen prompts. In auto-labeling workflows, such images can produce corrupted annotations used for downstream analysis or model training. This setting presents three challenges: defining targeted objectives over a prompt-dependent RGB space, optimizing them with output-only feedback, and transferring adversarial queries across prompts.

Our key insight is that VICL-based segmentation exhibits appearance-level regularities in its RGB outputs, particularly foreground--background contrast and instance-specific color assignments. Instead of matching a fully specified RGB target, we optimize objectives over these regularities. Color Hijack (CH) drives foreground colors toward the background appearance, suppressing foreground visibility. Task Hijack (TH) collapses instance-specific colors toward a shared assignment, disrupting instance distinctions after palette-based decoding. These property-level objectives avoid specifying a high-dimensional, prompt-specific RGB target, making them more suitable for output-only optimization and transfer across prompts.

A second insight is that cross-prompt transfer depends on the surrogate prompt used during optimization. Motivated by prior work on visual prompt selection~\cite{RethinkingandImproving}, we propose a Self-Prompt Surrogate (SPS) formed from the clean query and the model's corresponding clean output. Unlike an arbitrary external demonstration, SPS is aligned with the query being attacked and reflects the model's initial behavior on that image without requiring access to deployment prompts. This construction reduces dependence on prompt-specific content, palettes, and prompt compositions and improves transfer to unseen prompts. The complete workflow, from SPS construction and query-only optimization to CH and TH transfer under unseen deployment prompts, is illustrated in Figure~\ref{fig:teaser}.

We evaluate CH and TH on COCO~\cite{lin2015microsoftCOCO} and HierText~\cite{long2022towards} across four VICL models and multiple segmentation settings. Each adversarial query is optimized under a surrogate prompt and evaluated using held-out deployment prompts unavailable during optimization. We measure CH foreground suppression and TH color collapse, and further evaluate TH-induced changes in instance count, mask retention, and instance separation after palette-based decoding. We compare SPS with RPS under a paired protocol and assess practical impact through a VICL-based auto-labeling pipeline, where CH-corrupted annotations substantially degrade downstream model performance. Our contributions are:

\begin{itemize}
\item We formalize a query-only, prompt-agnostic threat model for VICL-based segmentation, where the attacker observes only returned RGB outputs and has no access to model internals or deployment prompts.

\item We identify the prompt-conditioned RGB output space as a distinct attack surface and introduce CH, which suppresses foreground visibility by driving foreground colors toward the background appearance, and TH, which disrupts instance distinctions by collapsing instance-specific colors toward a shared color assignment.

\item We propose SPS, which constructs an attack-time visual demonstration from the clean query and its corresponding clean output, and compare it with RPS under a paired protocol that controls for initialization and reference construction.

\item We evaluate the attacks across four VICL models, multiple segmentation settings, two datasets, and held-out deployment prompts, and demonstrate practical impact through TH-induced structural degradation after palette-based decoding and downstream training with CH-corrupted annotations.
\end{itemize}

\section{Background}
Some VICL methods~\cite{VisualPrompting2022, bai2024sequential} formulate visual in-context learning as discrete visual-token prediction using a VQ-GAN codebook. Although this provides a unified token space, discretization introduces quantization error and limits fine-grained prediction. Prior studies further report poor reconstruction by Visual Prompting even when the prompt contains the query image and its ground-truth output~\cite{stable-diffusion}, as well as difficulty distinguishing instances within the same semantic class with MAE-VQGAN~\cite{RethinkingandImproving}. Other methods instead represent task outputs directly as RGB images and learn continuous pixel-to-pixel prediction from visual demonstrations. Painter~\cite{Wang_Painter_CVPR} unifies multiple vision tasks through image-valued outputs and masked image modeling over concatenated input-output pairs. SegGPT~\cite{Wang_2023_ICCV} casts semantic, instance, and panoptic segmentation as in-context coloring, while ConText and ConTextV~\cite{zhang2025contextdrivingincontextlearning} extend this formulation to fine-grained text segmentation and removal. Related image-valued VICL methods also apply this paradigm to low-level vision and image restoration~\cite{rajagopalan2024awracle, liu2024PromptGIP}. Visual prompt selection and fusion studies further show that the choice and composition of demonstrations substantially affect VICL performance~\cite{NEURIPS2023_398ae57e, Wang25_Condenser}, especially for segmentation~\cite{RethinkingandImproving}.

This paper focuses on continuous, image-valued VICL segmentation models whose demonstrations specify the RGB output convention. In auto-labeling workflows, their prompt-conditioned RGB predictions are post-processed into structured annotations for downstream analysis or model training~\cite{labellerr2023seggpt,autodistill2023seggpt,encord2023seggpt}.

\section{Related Work}
Existing targeted black-box attacks on segmentation and related dense prediction broadly address conventional task-specific models~\cite{cai2023ensemble} and SAM-style promptable models~\cite{kirillov2023,zhang2023attacksamattackingsegmentmodel}. Unlike classification attacks, they must manipulate high-dimensional structured outputs, including dense label maps, instance masks, and prompt-conditioned masks, requiring coordinated pixel-level changes even for local targets. These challenges have also been documented in white-box settings~\cite{arnab_cvpr_2018,rony2022proximal}. 

For conventional semantic and instance segmentation, targeted objectives make selected pixels, regions, or instances match attacker-specified labels or masks. Early work demonstrates transferable pixel-level attacks, while later studies show that targeted transfer for structured outputs is substantially harder than for classification and often benefits from surrogate ensembles~\cite{A11_xie2017adversarial,cai2023ensemble}. These black-box methods therefore generally assume a fixed label or mask prediction space and optimize adversarial examples on accessible surrogate models before transferring them to the victim. Cai et al.~\citeyearpar{cai2023ensemble} also query the victim model to tune the surrogate ensemble weights. For SAM-style models~\cite{kirillov2023}, attacks manipulate object masks conditioned on spatial prompts such as points, boxes, or masks~\cite{lu2024unsegment}. Some optimize perturbations for fixed prompt configurations~\cite{zhang2023attacksamattackingsegmentmodel}, while others reduce prompt dependence by perturbing prompt-shared image representations~\cite{10.1109/TMM.2024.3521769}, aligning adversarial features with deformed targets~\cite{lu2024unsegment}, or defining region-level objectives effective under different clicks~\cite{10677988}. Despite this broader prompt coverage, these methods still target mask outputs and commonly rely on transferable features or surrogate models.

VICL-based segmentation differs from both settings. Its prompts are independent image--output demonstrations that define task semantics and RGB rendering conventions rather than spatial cues on the query, while its outputs are prompt-conditioned RGB predictions rather than fixed labels or masks. Because practical deployments may expose neither model internals nor the deployment prompt, existing label- or mask-space transfer attacks do not directly apply. We instead study query-only targeted attacks that use only returned RGB outputs to optimize appearance-space objectives that transfer across unseen deployment prompts.

\section{Attack Method}
\label{sec:method}

We formulate the attack over the prompt-conditioned RGB output space and target appearance-level properties, such as foreground--background contrast and instance-specific color assignments, rather than a fully specified pixel-aligned rendering.

Given a normalized query image \(x\in[0,1]^{H\times W\times 3}\), the attacker seeks a bounded perturbation \(\delta\) such that
\[
    x^{\mathrm{adv}}
    =
    \operatorname{clip}_{[0,1]}(x+\delta),
    \qquad
    \|\delta\|_{\infty}\leq\epsilon,
\]
and the returned RGB prediction exhibits a targeted appearance-level effect. Let \(f(\cdot;P)\) denote the black-box VICL system conditioned on visual prompt \(P\), and define
\[
    y_{\delta}^{P}=f(x^{\mathrm{adv}};P).
\]
During optimization, the attacker selects a surrogate prompt \(P_{\mathrm{sur}}\) and observes only the returned RGB outputs \(y_{\delta}^{P_{\mathrm{sur}}}\). The deployment prompt \(P_{\mathrm{dep}}\) remains unknown, and the attacker has no access to model parameters, logits, gradients, or intermediate features. The objective is therefore to optimize \(\delta\) using feedback under \(P_{\mathrm{sur}}\) such that the intended effect transfers to the unknown deployment prompt \(P_{\mathrm{dep}}\).

Our framework combines three components: SPS for query-aligned surrogate construction, the CH and TH appearance-space objectives, and a query-only zeroth-order optimizer for dense RGB outputs.

\subsection{Self-Prompt Surrogate}
\label{sec:sps}

Optimizing with an arbitrary external prompt can cause the perturbation to overfit to its image content, palette, or annotation layout. We instead construct the surrogate prompt directly from the query being attacked. Given an attacker-controlled initialization prompt \(P_{\mathrm{init}}\), the attacker first submits the clean query and receives
\[
y^0=f(x;P_{\mathrm{init}}).
\]
SPS then forms the query-aligned surrogate prompt
\[
P_{\mathrm{SPS}}=(x,y^0).
\]
This construction requires no access to \(P_{\mathrm{dep}}\) and ties the surrogate directly to the query and its returned clean output. For our proposed attack, we set \(P_{\mathrm{sur}}=P_{\mathrm{SPS}}\). RPS instead uses a randomly selected external demonstration \(P_{\mathrm{RPS}}\) as both \(P_{\mathrm{init}}\) and \(P_{\mathrm{sur}}\). Reference regions and colors are always extracted from the clean output obtained under the corresponding initialization prompt. In the paired comparison, the same external prompt \(P^{(r)}\) serves as \(P_{\mathrm{init}}\) for both methods, producing a shared clean output \(y^{0,(r)}=f(x;P^{(r)})\). RPS optimizes under \(P_{\mathrm{sur}}=P^{(r)}\), whereas SPS optimizes under \(P_{\mathrm{sur}}=(x,y^{0,(r)})\). The paired runs therefore differ only in surrogate construction.

\subsection{RGB Appearance-Space Objectives}
\label{sec:rgb_targets}

We extract all foreground pixels from \(y^0\) using the same foreground--background separation rule employed in output post-processing, without ground-truth annotations. Let \(R_f\) denote the resulting foreground region. We dilate \(R_f\) to include foreground boundaries and accommodate small spatial discrepancies in the parsed RGB output:
\[
    R_o
    =
    \operatorname{Dilate}(R_f),
    \qquad
    M_o
    =
    \mathbf{1}_{R_o},
\]
where \(R_o\) is the optimization ROI and \(M_o\) is the binary spatial mask used to restrict the input perturbation to \(R_o\). We use \(u\) to index pixel locations, and all RGB values are normalized to \([0,1]^3\).

\paragraph{Color Hijack.}
CH targets foreground--background contrast. Let \(c_{\mathrm{bg}}\in[0,1]^3\) denote the background color parsed from \(y^0\). CH minimizes
\[
    \mathcal{L}_{\mathrm{CH}}
    =
    \frac{1}{3|R_o|}
    \sum_{u\in R_o}
    \left\|
    y_{\delta,u}^{P_{\mathrm{sur}}}
    -
    c_{\mathrm{bg}}
    \right\|_2^2.
\]
This objective drives the attacked output within \(R_o\) toward the background appearance, suppressing foreground visibility.

\paragraph{Task Hijack.}
TH targets class-agnostic instance-style segmentation, in which distinct RGB assignments encode separate instances. Rather than suppressing the foreground, TH collapses these assignments toward a shared color. To reduce foreground--background color mixing near foreground boundaries, we define an eroded foreground interior
\[
R_c=\operatorname{Erode}(R_f)\subseteq R_f\subseteq R_o.
\]
We construct the shared target color from the mean clean RGB value within this interior:
\[
c^\star=\frac{1}{|R_c|}\sum_{u\in R_c}y^0_u.
\]
We further define the mean attacked color over the full ROI as
\[
    \bar y_{R_o}^{\delta}
    =
    \frac{1}{|R_o|}
    \sum_{u\in R_o}
    y_{\delta,u}^{P_{\mathrm{sur}}}.
\]
The TH objective is
\[
\begin{aligned}
    \mathcal{L}_{\mathrm{TH}}
    =
    &\;
    \frac{1}{|R_c|}
    \sum_{u\in R_c}
    \left\|
    y_{\delta,u}^{P_{\mathrm{sur}}}
    -
    c^\star
    \right\|_2^2
    \\[2pt]
    &+
    \frac{\lambda_{\mathrm{var}}}{3|R_o|}
    \sum_{u\in R_o}
    \left\|
    y_{\delta,u}^{P_{\mathrm{sur}}}
    -
    \bar y_{R_o}^{\delta}
    \right\|_2^2.
\end{aligned}
\]
The first term uses the eroded foreground interior to provide a stable color anchor that is less affected by foreground--background mixing. The second penalizes color variance over the full ROI, extending the collapse to foreground boundaries and surrounding ROI pixels.

\subsection{Query-Only RGB Optimization}
\label{sec:query_optimization}

Directly optimizing a full-resolution perturbation is inefficient using only returned RGB outputs. We instead optimize a low-dimensional RGB parameter \(p\), bilinearly upsample it to the model input resolution, restrict it to the optimization ROI, and project it onto the permitted \(L_\infty\) perturbation set:
\[
    \delta(p)
    =
    \Pi_{\epsilon}
    \left(
    M_o
    \odot
    \operatorname{Up}(p)
    \right),
\]
where \(M_o\) is broadcast across the RGB channels,
\(\operatorname{Up}(\cdot)\) denotes bilinear upsampling, and
\[
    \Pi_{\epsilon}(z)
    =
    \operatorname{clip}
    \left(
    z,-\epsilon,\epsilon
    \right)
\]
denotes element-wise projection onto the \(L_\infty\) perturbation set.

We optimize \(p\) using a standard simultaneous-perturbation
zeroth-order estimator~\cite{spall1992multivariate} in the
low-dimensional, ROI-constrained perturbation space defined above.
At iteration \(t\), let \(p_t\) denote the current low-dimensional
perturbation parameter and let \(\mu_t>0\) denote the finite-difference
radius. For independent Rademacher directions
\[
    v_k
    \in
    \{-1,+1\}^{\dim(p)},
    \qquad
    k=1,\ldots,K,
\]
we estimate
\[
    \hat g_t
    =
    \frac{1}{K}
    \sum_{k=1}^{K}
    \frac{
    \mathcal{L}(p_t+\mu_t v_k)
    -
    \mathcal{L}(p_t-\mu_t v_k)
    }{
    2\mu_t
    }
    v_k,
\]
where \(\mathcal{L}(p)\) denotes either the CH or TH objective evaluated on
the RGB output induced by \(\delta(p)\). The symmetric estimator requires
\(2K\) model evaluations, followed by one additional evaluation of the
updated iterate for objective tracking and best-iterate selection.

We update \(p\) using the mean-normalized estimated gradient:
\[
    p_{t+1}
    =
    p_t
    -
    \eta_t
    \frac{\hat g_t}{
    \max\!\left(
    \operatorname{mean}(|\hat g_t|),
    \tau
    \right)},
\]
where \(\eta_t\) is the step size and \(\tau>0\) is a small constant for
numerical stability. All objective evaluations use only RGB outputs returned
under the surrogate prompt.

The final adversarial iterate is selected using only the surrogate objective,
without querying \(P_{\mathrm{dep}}\). To account for image serialization in
practical labeling pipelines, the selected adversarial image is saved as a PNG and reloaded before evaluation.

\section{Attack Evaluation}
\label{sec:attack_evaluation}

We evaluate CH across four VICL models with semantic-, instance-, and text-style RGB outputs, and TH on instance-style outputs. We compare SPS with RPS under a paired initialization protocol, compare our objectives with generic attack baselines, analyze query-budget effects, test transfer across palettes, output styles, and prompt compositions, and examine the practical impact of TH and CH.

\subsection{Experimental Setup}

\paragraph{Models and data.}
We evaluate Painter~\cite{Wang_Painter_CVPR} and SegGPT~\cite{Wang_2023_ICCV} on the same 200 query images drawn from the COCO person-keypoint subset~\cite{lin2015microsoftCOCO}, each containing at least two clearly visible, annotated person instances. All COCO surrogate prompts and held-out deployment prompts are drawn from disjoint images in the same subset. We evaluate ConText and ConTextV~\cite{zhang2025contextdrivingincontextlearning} on 50 HierText query images~\cite{long2022towards}. CH is evaluated on semantic- and instance-style outputs from Painter and SegGPT and text-style outputs from ConText and ConTextV. TH is evaluated on instance-style Painter and SegGPT outputs.

\paragraph{Attack configuration.}
We optimize \(p\in\mathbb{R}^{64\times64\times3}\) and bilinearly upsample it to the \(448\times448\) query resolution. The optimization ROI \(R_o\) is constructed by dilating the parsed foreground \(R_f\) with a \(25\times25\) kernel. For TH, the interior region \(R_c\) is obtained by eroding \(R_f\) with a \(5\times5\) kernel. Each run uses 600 iterations, \(K=32\) Rademacher directions per iteration, and the mean-normalized update. For CH, we use \((\mu,\eta)=(0.031373,0.007843)\). For TH, we set \(\lambda_{\mathrm{var}}=0.12\) and use \((\mu,\eta)\) values of \((0.03137,0.00784)\), \((0.01569,0.00294)\), and \((0.00784,0.00147)\), beginning at iterations \(0\), \(250\), and \(450\), respectively. Unless otherwise specified, \(\epsilon\in\{8/255,16/255,32/255\}\).

\paragraph{Prompt protocol.}
For the main SPS evaluations, each model--style setting uses one fixed, task-matched non-query image--output pair as \(P_{\mathrm{init}}\) for all attack queries. This pair is fixed before evaluation, disjoint from the attack queries and held-out deployment prompts, and used only to obtain \(y^0\); optimization then uses the query-specific \(P_{\mathrm{SPS}}=(x,y^0)\).

Each model--style setting has ten task-matched held-out deployment pairs, drawn from the COCO person-keypoint subset for Painter and SegGPT and from HierText for ConText and ConTextV. These pairs are disjoint from the attack queries and unavailable during optimization. Each adversarial query is evaluated under all ten prompts, and the results are averaged within the query. The same held-out prompts are used across methods, budgets, and ablations.

For the paired SPS--RPS comparison, each query samples two task-matched non-query prompts, \(P^{(1)}\) and \(P^{(2)}\), from a pool disjoint from the attack queries and held-out deployment prompts. The pool follows the same data source as the queries; COCO prompts use disjoint images from the same person-keypoint subset, so RPS is matched to the attack queries in dataset, task, output style, and the presence of annotated persons.

For each \(r\in\{1,2\}\), we obtain \(y^{0,(r)}=f(x;P^{(r)})\). RPS uses \(P_{\mathrm{init}}=P_{\mathrm{sur}}=P^{(r)}\), whereas SPS uses \(P_{\mathrm{sur}}=(x,y^{0,(r)})\). The pair shares the clean output, references, optimization settings, budget, and deployment prompts. We average first over deployment prompts and then over the two initializations.

\paragraph{Metrics.}
Metrics are computed for each query--prompt output, averaged across deployment prompts within each query, and then aggregated across queries. The \(\pm\) values denote standard deviations across queries; otherwise, entries are means unless explicitly labeled as medians. For semantic- and text-style outputs, foreground regions are extracted through channel-wise discretization according to the corresponding RGB label encoding. For instance-style outputs, we use the corresponding color-kernel post-processing procedure~\cite{Wang_Painter_CVPR}.

\begin{figure}[!htb]
\centering
\includegraphics[width=0.74\columnwidth]{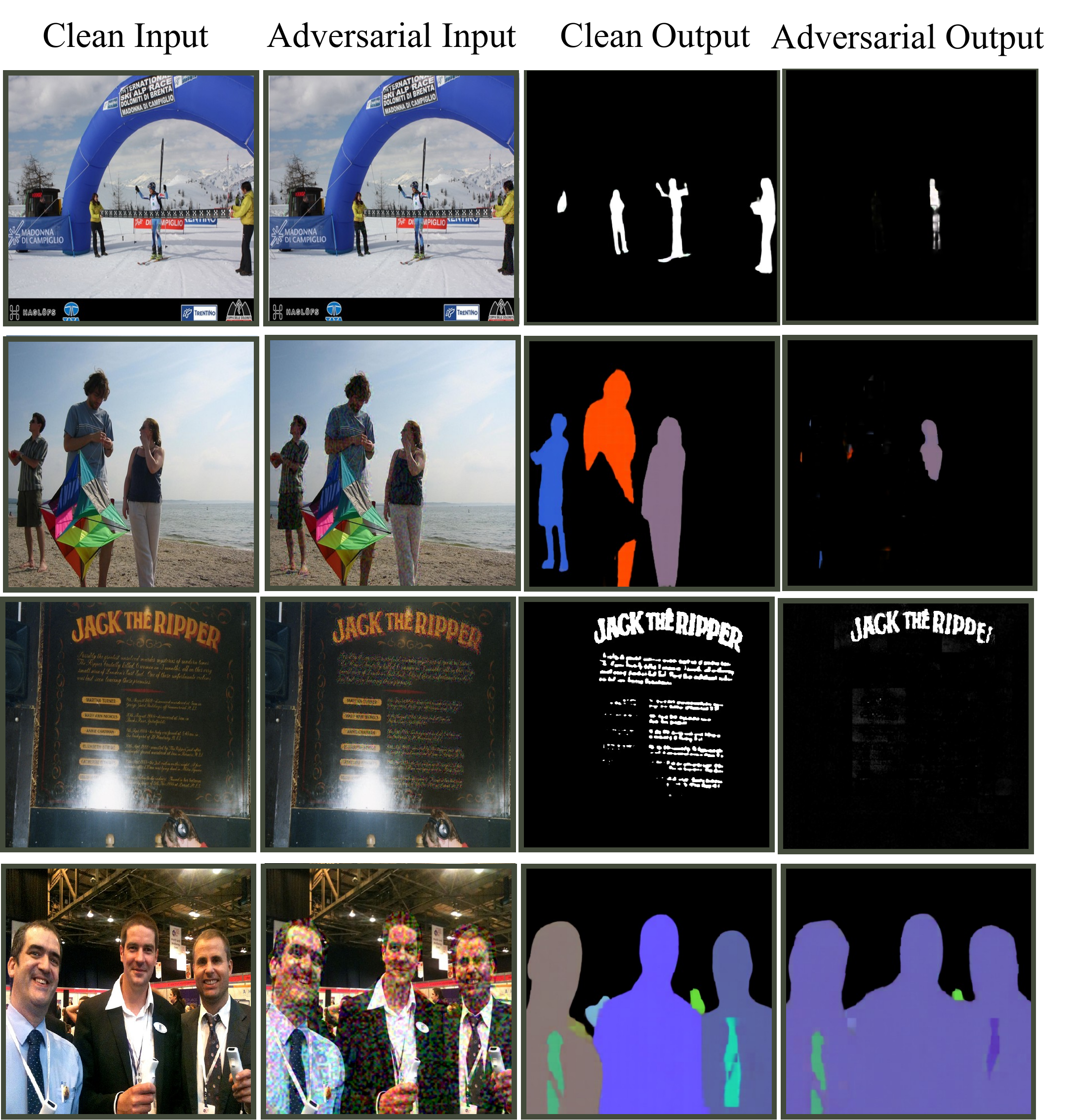}
\caption{Qualitative CH and TH results. The first three rows show semantic-, instance-, and text-style CH; the last row shows instance-style TH.}
\label{fig:qual_ch_th}
\end{figure}

For CH, we compute foreground retention between clean and attacked outputs under the same deployment prompt:
\[
\operatorname{Retain}=\frac{|F_{\mathrm{clean}}\cap F_{\mathrm{adv}}|}{|F_{\mathrm{clean}}|},
\]
where lower values indicate stronger foreground suppression.

For TH, let \(R_{\mathrm{eval}}\) be the foreground parsed from the clean output and \(z_u\) the attacked RGB value at \(u\in R_{\mathrm{eval}}\). We fit a \(k\)-means palette with \(k=8\) to the clean RGB values and assign each attacked pixel to its nearest clean palette center. Let \(c_{\mathrm{dom}}\) be the center receiving the most assignments. We compute
\[
\operatorname{TolDist}_{\mathrm{pal}}=\frac{1}{|R_{\mathrm{eval}}|}\sum_{u\in R_{\mathrm{eval}}}\left[\|z_u-c_{\mathrm{dom}}\|_2-\rho\right]_+,
\]
where \(\rho=12/255\), and
\(
\operatorname{CC}=\sum_i\pi_i^2,
\)
where \(\pi_i\) is the fraction of attacked pixels assigned to the \(i\)-th palette center. Stronger instance-color collapse yields lower \(\operatorname{TolDist}_{\mathrm{pal}}\) and higher CC.

To measure decoded structural effects, we apply a palette-based instance decoder to the clean and attacked outputs. Count Ratio is the ratio of attacked to clean instance counts. Mask Ret.@50 is the fraction of clean instances whose maximum IoU with any attacked instance is at least \(0.5\), without one-to-one matching. For Pair Merge, each clean instance is assigned to the attacked instance with which it has the largest positive intersection and remains unassigned when no positive overlap exists; the metric is the fraction of clean-instance pairs assigned to the same attacked instance. Lower Count Ratio and Mask Ret.@50 and higher Pair Merge indicate stronger structural degradation.

\begin{figure}[!htb]
\centering
\includegraphics[width=0.9\columnwidth]{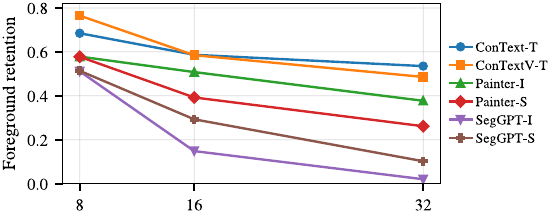}
\caption{CH foreground retention under held-out prompts across \(\epsilon\in\{8,16,32\}/255\). Lower is stronger; S, I, and T denote semantic-, instance-, and text-style outputs.}
\label{fig:ch_budget}
\end{figure}

\subsection{Main Results}

Figure~\ref{fig:qual_ch_th} illustrates CH foreground suppression and TH instance-color collapse. Figure~\ref{fig:ch_budget} further shows that CH transfers across held-out semantic-, instance-, and text-style prompts, with the strongest suppression on SegGPT.

\paragraph{Task Hijack.}
Table~\ref{tab:th_budget}(a) shows instance-color collapse on Painter and SegGPT. SegGPT largely saturates after \(16/255\), while Painter exhibits increasing color concentration (CC) despite similar tolerance distances. Table~\ref{tab:th_budget}(b) shows that adding the variance term improves the mean metrics, with \(\lambda_{\mathrm{var}}=0.12\) giving the best mean values.

\begin{table}[t]
\centering
\small
\setlength{\tabcolsep}{1.5pt}

\textbf{(a) TH across perturbation budgets}

\begin{tabular*}{\columnwidth}{@{\extracolsep{\fill}}llcc@{}}
\toprule
Model & \(\epsilon\) & Pal. TolDist & CC\\
\midrule
Painter
 & Clean      & \(0.315 \pm 0.123\) & \(0.269 \pm 0.121\)\\
 & \(8/255\)  & \(0.228 \pm 0.134\) & \(0.440 \pm 0.205\)\\
 & \(16/255\) & \(0.233 \pm 0.139\) & \(0.463 \pm 0.216\)\\
 & \(32/255\) & \(0.227 \pm 0.131\) & \(0.483 \pm 0.226\)\\
SegGPT
 & Clean      & \(0.344 \pm 0.148\) & \(0.327 \pm 0.128\)\\
 & \(8/255\)  & \(0.218 \pm 0.140\) & \(0.467 \pm 0.233\)\\
 & \(16/255\) & \(0.181 \pm 0.134\) & \(0.533 \pm 0.271\)\\
 & \(32/255\) & \(0.176 \pm 0.126\) & \(0.534 \pm 0.276\)\\
\bottomrule
\end{tabular*}

\smallskip
\textbf{(b) TH variance-weight ablation}

\begin{tabular*}{\columnwidth}{@{\extracolsep{\fill}}ccc@{}}
\toprule
\(\lambda_{\mathrm{var}}\) & Pal. TolDist & CC\\
\midrule
0    & \(0.187 \pm 0.131\) & \(0.513 \pm 0.260\)\\
0.05 & \(0.183 \pm 0.134\) & \(0.529 \pm 0.270\)\\
0.12 & \(0.181 \pm 0.134\) & \(0.533 \pm 0.271\)\\
\bottomrule
\end{tabular*}

\caption{TH results and objective ablation. Attacks use the
default SPS and held-out prompts. Panel (b) uses SegGPT at
\(\epsilon=16/255\).}
\label{tab:th_budget}
\end{table}

\paragraph{Paired SPS--RPS comparison.}
Table~\ref{tab:sps_rps} pairs initialization prompts, clean outputs, references, optimization settings, budgets, and deployment prompts, isolating surrogate construction. RPS is task- and domain-matched; its COCO prompts come from the same person-keypoint subset as the attack queries. Against this matched baseline, SPS reduces SegGPT instance/semantic CH retention from 0.918/0.792 to 0.155/0.364 and TH TolDist$_{\mathrm{pal}}$ from 0.345 to 0.225.

\begin{table}[t]
\centering
\small
\setlength{\tabcolsep}{1.5pt}

\textbf{(a) CH foreground retention}

\begin{tabular*}{\columnwidth}{@{\extracolsep{\fill}}lcc@{}}
\toprule
Model and style & SPS & RPS\\
\midrule
Painter, instance
  & \(0.460 \pm 0.219\) & \(0.627 \pm 0.344\)\\
Painter, semantic
  & \(0.293 \pm 0.203\) & \(0.404 \pm 0.308\)\\
SegGPT, instance
  & \(0.155 \pm 0.133\) & \(0.918 \pm 0.196\)\\
SegGPT, semantic
  & \(0.364 \pm 0.228\) & \(0.792 \pm 0.357\)\\
ConText, text
  & \(0.522 \pm 0.208\) & \(0.784 \pm 0.267\)\\
ConTextV, text
  & \(0.613 \pm 0.211\) & \(0.833 \pm 0.182\)\\
\bottomrule
\end{tabular*}

\smallskip
\textbf{(b) TH palette-aware tolerance distance}

\begin{tabular*}{\columnwidth}{@{\extracolsep{\fill}}lcc@{}}
\toprule
Model & SPS & RPS\\
\midrule
Painter & \(0.244 \pm 0.126\) & \(0.278 \pm 0.140\)\\
SegGPT & \(0.225 \pm 0.135\) & \(0.345 \pm 0.152\)\\
\bottomrule
\end{tabular*}

\caption{Paired SPS--RPS comparison at
\(\epsilon=16/255\). Runs differ only in surrogate construction
and are averaged over two initializations per query. Lower is
better.}
\label{tab:sps_rps}
\end{table}

\paragraph{Comparison with generic attack baselines.} We compare against random ROI perturbations, untargeted output disruption, and pixel-aligned RGB matching. Random perturbations are sampled in the same low-dimensional ROI and averaged over five draws. Output disruption minimizes \(\mathcal{L}_{\mathrm{OD}}=-\frac{1}{3|R_o|}\sum_{u\in R_o}\|y^{P_{\mathrm{SPS}}}_{\delta,u}-\tilde{y}^{0}_{u}\|_2^2\), where \(\tilde{y}^{0}=f(x;P_{\mathrm{SPS}})\). RGB matching is given the exact surrogate output produced by the selected CH or TH example and is re-optimized from scratch. All optimized methods share the same SPS, ROI, perturbation and query budgets, and held-out prompts. Table~\ref{tab:rgb_baselines} shows that only our objectives produce strong CH and TH effects.

\begin{table}[t]
\centering
\small
\setlength{\tabcolsep}{1pt}

\begin{tabular*}{\columnwidth}{@{\extracolsep{\fill}}lccc@{}}
\toprule
Method
& CH-I Ret.
& CH-S Ret.
& TH Pal. TolDist\\
\midrule
Clean output
& 1.000 & 1.000 & 0.344\\
Random ROI
& 0.987 & 0.988 & 0.345\\
Output disruption
& 0.997 & 0.992 & 0.397\\
RGB matching
& 0.999 & 0.992 & 0.344\\
CH / TH (ours)
& \textbf{0.148}
& \textbf{0.293}
& \textbf{0.181}\\
\bottomrule
\end{tabular*}

\caption{Generic attack baselines on SegGPT at \(\epsilon=16/255\). Lower values indicate stronger attacks. CH-I and CH-S denote instance- and semantic-style outputs.}
\label{tab:rgb_baselines}
\end{table}

\begin{table}[t]
\centering
\small
\setlength{\tabcolsep}{0.5pt}

\textbf{(a) CH foreground retention}

\begin{tabular*}{\columnwidth}{@{\extracolsep{\fill}}lccccc@{}}
\toprule
& & \multicolumn{2}{c}{Instance}
  & \multicolumn{2}{c}{Semantic}\\
\cmidrule(lr){3-4}\cmidrule(lr){5-6}
Steps & Eval. & Mean & Med. & Mean & Med.\\
\midrule
100 & \(6.5\)k  & 0.651 & 0.824 & 0.623 & 0.721\\
200 & \(13.0\)k & 0.508 & 0.591 & 0.497 & 0.554\\
400 & \(26.0\)k & 0.265 & 0.226 & 0.368 & 0.414\\
600 & \(39.0\)k & 0.148 & 0.002 & 0.293 & 0.297\\
\bottomrule
\end{tabular*}

\smallskip
\textbf{(b) TH instance-color collapse}

\begin{tabular*}{\columnwidth}{@{\extracolsep{\fill}}lccccc@{}}
\toprule
& & \multicolumn{2}{c}{Pal. TolDist}
  & \multicolumn{2}{c}{CC}\\
\cmidrule(lr){3-4}\cmidrule(lr){5-6}
Steps & Eval. & Mean & Med. & Mean & Med.\\
\midrule
100 & \(6.5\)k  & 0.257 & 0.277 & 0.427 & 0.369\\
200 & \(13.0\)k & 0.210 & 0.224 & 0.491 & 0.426\\
400 & \(26.0\)k & 0.184 & 0.163 & 0.506 & 0.469\\
600 & \(39.0\)k & 0.179 & 0.158 & 0.538 & 0.460\\
\bottomrule
\end{tabular*}

\caption{Query-budget ablation on SegGPT at
\(\epsilon=16/255\). Evaluation counts exclude initialization.}
\label{tab:query_budget}
\end{table}

\paragraph{Query-budget ablation.}
Table~\ref{tab:query_budget} shows that both attacks are effective at intermediate budgets. At 200 iterations (\(13.0\)k evaluations), instance- and semantic-style CH retention reaches \(0.508\) and \(0.497\), while TH \(\operatorname{TolDist}_{\mathrm{pal}}\) reaches \(0.210\). TH approaches its final performance by 400 iterations, whereas CH continues to improve through 600 iterations. The 600-step TH values differ slightly from the main results because this ablation evaluates fixed checkpoints rather than the surrogate-selected best iterate.

\subsection{Robustness and Generalization}

\paragraph{Cross-palette and cross-style transfer.}
Alternative-palette prompts preserve the output style while using different RGB label colors. Table~\ref{tab:palette} shows that default-SPS attacks transfer across palettes and between semantic- and instance-style prompts.

\paragraph{Few-shot prompts.}
Without re-optimization, we evaluate default-SPS adversarial
queries under all 1-, 2-, and 3-shot compositions of the ten
held-out pairs (9 two-shot and 36 three-shot combinations per
anchor). Table~\ref{tab:fewshot} reports the averages.

\begin{table}[t]
\centering
\small
\setlength{\tabcolsep}{1.5pt}

\textbf{(a) Cross-palette CH retention}

\begin{tabular*}{\columnwidth}{@{\extracolsep{\fill}}lcc@{}}
\toprule
Evaluation & Instance & Semantic\\
\midrule
Original palette
  & \(0.148 \pm 0.206\) & \(0.293 \pm 0.289\)\\
Alternative palette
  & \(0.355 \pm 0.282\) & \(0.322 \pm 0.289\)\\
\bottomrule
\end{tabular*}

\smallskip
\textbf{(b) Cross-style CH retention}

\begin{tabular*}{\columnwidth}{@{\extracolsep{\fill}}lc@{}}
\toprule
Optimization \(\rightarrow\) evaluation & Retain\\
\midrule
Semantic \(\rightarrow\) instance
  & \(0.533 \pm 0.324\)\\
Instance \(\rightarrow\) semantic
  & \(0.532 \pm 0.340\)\\
\bottomrule
\end{tabular*}

\smallskip
\textbf{(c) Cross-palette TH color collapse}

\begin{tabular*}{\columnwidth}{@{\extracolsep{\fill}}lcc@{}}
\toprule
Evaluation & Pal. TolDist & CC\\
\midrule
Original clean
  & \(0.344 \pm 0.148\) & \(0.327 \pm 0.128\)\\
Original attacked
  & \(0.181 \pm 0.134\) & \(0.533 \pm 0.271\)\\
Alternative clean
  & \(0.318 \pm 0.150\) & \(0.325 \pm 0.127\)\\
Alternative attacked
  & \(0.242 \pm 0.144\) & \(0.394 \pm 0.135\)\\
\bottomrule
\end{tabular*}

\caption{Cross-palette and cross-style transfer on SegGPT at
\(\epsilon=16/255\).}
\label{tab:palette}
\end{table}

\begin{table}[t]
\centering
\small
\setlength{\tabcolsep}{1pt}

\textbf{(a) CH foreground retention}

\begin{tabular*}{\columnwidth}{@{\extracolsep{\fill}}lcc@{}}
\toprule
Setting & Instance & Semantic\\
\midrule
1-shot & \(0.148 \pm 0.206\) & \(0.293 \pm 0.289\)\\
2-shot & \(0.125 \pm 0.177\) & \(0.317 \pm 0.291\)\\
3-shot & \(0.108 \pm 0.174\) & \(0.333 \pm 0.294\)\\
\bottomrule
\end{tabular*}

\smallskip
\textbf{(b) TH instance-color collapse}

\begin{tabular*}{\columnwidth}{@{\extracolsep{\fill}}lcccc@{}}
\toprule
& \multicolumn{2}{c}{Pal. TolDist}
& \multicolumn{2}{c}{CC}\\
\cmidrule(lr){2-3}\cmidrule(lr){4-5}
Setting & Clean & Att. & Clean & Att.\\
\midrule
1-shot & 0.344 & 0.181 & 0.327 & 0.533\\
2-shot & 0.299 & 0.191 & 0.312 & 0.528\\
3-shot & 0.246 & 0.192 & 0.281 & 0.575\\
\bottomrule
\end{tabular*}

\caption{Few-shot transfer on SegGPT at
\(\epsilon=16/255\).}
\label{tab:fewshot}
\end{table}

\subsection{Practical Impact}
\label{sec:practical_impact}

\paragraph{Decoded instance structure.}
Table~\ref{tab:th_structure} shows that TH reduces decoded instance counts and mask retention while inducing many-to-one instance merges.

\begin{table}[t]
\centering
\small
\setlength{\tabcolsep}{1.5pt}

\begin{tabular*}{\columnwidth}{@{\extracolsep{\fill}}lccc@{}}
\toprule
Model & Count Ratio & Mask Ret.@50 & Pair Merge\\
\midrule
Painter & 0.733 & 0.058 & 0.205\\
SegGPT & 0.671 & 0.237 & 0.397\\
\bottomrule
\end{tabular*}

\caption{TH impact on palette-decoded structure at
\(\epsilon=16/255\). Attacks use the default SPS; values are means relative to clean outputs. Lower Count Ratio and Mask Ret.@50 and higher Pair Merge indicate stronger degradation.}
\label{tab:th_structure}
\end{table}

\paragraph{Downstream training impact.} 
Using Autodistill's public VICL pipeline~\cite{autodistill2023seggpt}, SegGPT serves as the upstream annotator and its outputs supervise a downstream YOLO-seg model. As CH corruption among 120 annotation inputs increases from \(0\%\) to \(30\%\), clean-validation mAP50 drops from \(0.761\) to \(0.281\).

\begin{figure}[!htb]
\centering
\includegraphics[width=0.8\columnwidth]{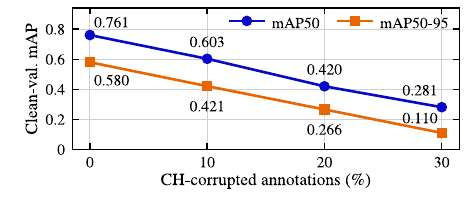}
\caption{Downstream impact of CH-corrupted annotations. mAP on 30 clean validation images.}
\label{fig:downstream}
\end{figure}
\section{Conclusion}
We presented query-only, prompt-agnostic targeted attacks on VICL-based segmentation. CH suppresses foregrounds, whereas TH collapses instance-specific colors; SPS constructs a query-aligned surrogate from returned RGB outputs. CH transfers across held-out prompts on four models and TH on Painter and SegGPT. On SegGPT, both attacks transfer across palettes and few-shot prompts, while CH also transfers across output styles. TH degrades decoded instance structure, and CH-corrupted annotations reduce downstream YOLO-seg mAP50 from 0.761 to 0.281.

Defenses must balance robustness with prompt adaptability and automation. Stronger prompt invariance may resist attacks but limit legitimate adaptation, while human review reduces scalability. Since the attacks transfer across unseen prompts and palettes, randomization alone is unlikely to suffice. Practical defenses could combine robust training, trusted-prompt consistency checks, and automated validation of RGB appearance and decoded instance structure, reserving human review for high-risk annotations.

\bibliography{reference}

\end{document}